# On the Military Applications of Large Language Models

Satu Johansson and Taneli Riihonen
*Tampere University*
Tampere, Finland
{satu.johansson, taneli.riihonen}@tuni.fi

*Abstract*—In this paper, military use cases or applications and implementation thereof are considered for natural language processing and large language models, which have broken into fame with the invention of the generative pre-trained transformer (GPT) and the extensive foundation model pretraining done by OpenAI for ChatGPT and others. First, we interrogate a GPT-based language model (viz. Microsoft Copilot) to make it reveal its own knowledge about their potential military applications and then critically assess the information. Second, we study how commercial cloud services (viz. Microsoft Azure) could be used readily to build such applications and assess which of them are feasible. We conclude that the summarization and generative properties of language models directly facilitate many applications at large and other features may find particular uses. This paper was originally presented at the NATO Science and Technology Organization Symposium (ICMCIS) organized by the Information Systems Technology (IST) Scientific and Technical Committee, IST-209-RSY – the ICMCIS, held in Oeiras, Portugal, 13-14 May 2025.

## I. Introduction

It is existentially important for nations and their armed forces to harness artificial intelligence (AI) and machine learning (ML) to security and defense applications. Within the wide discipline of ML/AI, our paper studies natural language processing (NLP) for written text. This subject has become extremely popular with the advent of language models that are based on the generative pre-trained transformer (GPT) introduced in [1], [2] and with the staggering pretraining effort done by companies such as OpenAI and Google that manifested foundational large language models (LLMs).

There is not yet much scientific literature on explicit military applications of LLMs or GPTs as opposed to military ML/AI in general, which involves many NLP concepts, too. The most relevant reference to our study is [3], where Biswas itemizes claims made by ChatGPT that it may have a role in

"military applications such as automated target recognition, military robotics, material development systems testing in simulation, military medicine, battle space autonomy, intelligence analysis, record tracking, military logistics, information warfare, driverless vehicles, surveillance, lethal autonomous weapons systems, battlefield environmental support, virtual and augmented reality modeling and simulations, free air combat modeling, missile guidance, communications, and network security, data fusion for situational awareness, swarm intelligence for swarm combat, autonomous flight control of UAVs [unmanned aerial vehicles], AI satellites and software-defined satellites, wearable systems for individual personnel, management of large amounts of military data and counter-AI operations." [3]

The listed blue-sky subjects seem too general to be useful in practice for more than gaining inspiration and also "the author states that … the actual possibilities currently are limited" [3]. Moreover, many of them are clearly tasks for ML/AI in general rather than any applications for GPT-based LLMs, or even NLP technology. Thus, language models' knowledge of prospective applications deserves another, closer inspection.

As for the other open literature, ML/AI and LLMs have been discussed in length in several publications for both military and related civilian contexts. [4] highlighted the emerging capabilities of AI in military use, especially in decision-making support, situational awareness, and management of massive data. [5] discussed ML learning strategies, namely reinforcement learning from human feedback (RLHF) and briefly LLMs for unmanned aircraft swarms. LLMs and ChatGPT as tools for real-time decision-making, especially in conjunction with autonomous driving, was discussed in [6]. A related military application, real-time control of LLM-based UAVs and several architectures thereof, was presented in [7]. Clearly a popular topic, ML techniques for UAVs were surveyed in [8]. Handling surveillance reports in the medical field with GPT and other generative LLMs was discussed in [9]. The reliability of ChatGPT for civilian medicine has been explored in the case study by [10].

Another very relevant reference to our study is [11], which discusses the general process of training and fine-tuning LLMs for tasks beyond NLP, including the capability of combining raw data from various sources and formats. The study then highlights several military domains where the LLMs' capabilities could potentially enhance present solutions, including air operations and assisting human operators in intelligence, surveillance, target acquisition, reconnaissance, and courses of action (COAs). As the challenges of LLMs are also considerable, including but not limited to the quality or cost of data storage or training and unwanted behaviors, like bias, hallucination, and lack of accountability, the author of [11] urges exploring the military potential of LLMs but as aids and enhancements to human operators, not as their replacements.

For more specific areas of defense research, [12] discussed unpredictable LLM behavior escalating conflicts in wargames and cautioned against trusting AI with real-life strategic military or diplomatic decision-making. The study in [13] discussed creating military conversational agents or chatbots with help of NLP and LLMs to provide situational information to soldiers in their native language, while the authors [14] employed LLMs to generate and analyze COAs for military operation. For a more situational approach, the study in [15] evaluated critically the deployment of AI-enabled decision support systems (DSS) from awareness to planning and prediction. The emerging capabilities of AI and GPT, including latency considerations, edge computing, and computational requirements especially for LLMs, along with several applications, were discussed at length in [16]. Related to information warfare, [17] used LLMs for detecting fake news and misinformation as well as for fact-checking.

The first main contribution of this paper is to characterize prospective military applications of GPT-based language models. Not so unlike [3] with ChatGPT, we asked Microsoft Copilot to explain the military applications of LLMs. However, unlike [3] that copied the LLM's generations without human consideration, we used those conversations only as research material and wrote this paper completely by ourselves with scientific judgment. Without forgetting the advances already presented in [4]–[17], we believe that our study serves as a starting point for innovating and developing applications beyond the state-of-the-art knowledge contained collectively in the language models' pretraining data, i.e., virtually all the text that the humankind has published so far.

The second main contribution of this paper is to describe how LLM/NLP technologies that are available 'commercially off the cloud' (COTC) — cf. 'commercially off the shelf' (COTS) — could be used readily for building military applications and assess the potential of use cases identified by the LLM. We present a case study on the feasibility and implementation by considering Microsoft Azure cloud services, in particular Azure OpenAI and AI Language, while it is reasonable to expect that comparable conclusions would be obtained with any other leading cloud platform, like Amazon Web Services (AWS) and Google Cloud, too. We conclude that, especially, the summarization and generative capabilities of LLMs could be directly adopted to build military applications, but there remains severe concern on accessibility and operations security in using cloud services. Likewise, also other NLP capabilities are directly available as components for military solutions.

The remainder of this paper is organized as follows. The next section overviews some ongoing research in the defense sector to complement this introduction. Section III discusses the characterization of military applications for LLMs based on Copilot's knowledge and our consideration, while Section IV discusses the capabilities of commercial-off-the-cloud AI models. A synthesis of these discussions is presented in Section V leading to a conclusion.

## II. Overview of Defense Technology Research

In this section, we survey research on the military applications of LLMs within NATO and its Science & Technology Organization (STO) based on public descriptions. In particular, there are two technical activities: The System Analysis & Studies (SAS) and Human Factors & Medicine (HFM) Panels' exploratory team (ET) SAS-HFM-ET-FM *"Natural Language Processing for Defence: Exploiting the Cutting Edge of Large Language Models for Military Contexts"* [18] as well as the Information Systems Technology (IST) Panel's research task group (RTG) IST-207 *"Military Applications for Large Language Models"* [19].

The recently completed SAS-HFM-ET-FM highlights the need to incorporate accurate semantic knowledge of military language and jargon, especially in conversational AI applications like chatbots. The activity first sought to ascertain and evaluate the existing industry-grade LLMs for military context. It then sought to develop and incorporate new scientific models for conversational AI and NLP applications in military use. The goal for their exploration has been to prepare a technical activity proposal for a possible follow-on RTG, but information about it has not been published yet at the STO website in January 2025. [18]

The ongoing IST-207 focuses on following the rapid technical advances of multimodal LLMs mainly in the private sector, their application through application programming interfaces (APIs), and application of proper training data and prompt engineering to best make use of the models' abilities. The goal is also to evaluate and establish military use cases by following the advances in the civilian sector while keeping in mind that civilian commercial models may exhibit behaviors especially unwanted in the military context, such as hallucination or political bias. [19]

Furthermore, the Internet is filled with non-scientific articles and blog texts suggesting that applications for LLM/NLP technologies are being developed by NATO member states round the globe. Reviewed open-source intelligence (OSINT) indicates the following: the military must develop its own LLMs rather than adopt commercial ones if generative AI is truly to be embraced [20], while addressing concerns over protecting sensitive information and preventing model tampering [21]. [22] cautions against hype and alarmist attitudes while acknowledging the LLMs' emerging capabilities, and [23] lists the pros and cons of LLMs in military use in equal measure. [24] deemed feasible some applications that our research considers to be outside the present capabilities of AI or better handled without LLMs.

## III. Prospective Military Applications

In this section, we report our investigation into a language model's knowledge of military applications for itself and its kind. To generate and collect the research material, we had several conversations with Microsoft Copilot, which is based on GPT-4 Turbo from OpenAI and pretrained with data that includes information up until October 2023. Starting from the high-level generic list presented in [3], we asked the model to elaborate on specific applications one by one in detail and covered all aspects as comprehensively as possible until we had exhausted all ideas on how to make it reveal any more information. We then studied the conversations carefully for compiling the following summary and characterization that should (more or less, we hope) represent the language model's contemporary knowledge of prospective military use cases.

The different fields on which the LLM was questioned could be divided into these rough categories: Summarizing existing materials, even massive data; generating new materials and simulations; providing analysis and predictions based on existing data; and surveillance and real-time reactions to perceived or potential threats. These categories manifest in the following sections, as the applications for LLMs for military use are discussed individually.

### A. Generating Training Materials and Summaries

For generating training materials and summaries of existing materials, Copilot was questioned on several points and tested on some. For military materials that would benefit from summarization, Copilot suggested field, training, and technical manuals, operational reports, intelligence briefings and mission briefs, policy documents, and standard operating procedures (SOPs)—in short, lengthy documents used in the field that can benefit from a concise presentation. When the conversation was repeated on another day, Copilot also suggested summarizing e-modules, field training exercises (FTX), after-action reviews (AARs), and even virtual reality and augmented reality (VR/AR) simulations. While interesting, the changes in answers for no apparent reason also highlight a major concern for any application of LLMs: the lack of consistency that would need to be addressed and the final solutions tested excessively for any military applications. Especially as LLMs are at least considered — if not already tested or implemented — for multiple critical fields, their inherently unpredictable behavior poses a significant hindrance to the adoption of AI.

Copilot was also able to give suggestions for making sure the outcome was as close to desired as possible. It suggested using annotated documents with the highlights and focal points included, in case its training data did not have examples of materials of this kind. At the same time, Copilot did claim some familiarity with military documents and suggested that some, publicly available materials had been used in the training of the model.

For benefits of entrusting an LLM with summarization tasks for military purposes, Copilot cited automating repetitive tasks, customizable content to engage the intended audience, saving time and resources, and ensuring a consistent and accessible style in the materials. However, on further questioning, it admitted that human oversight, expertise, and controlling the LLM's work was essential for quality control on most occasions, leaving humans still in the equation.

The answers provided by the LLM were mostly reasonable and well-rounded, and when asked to perform summarization tasks for a .docx-format handbook, a select handbook passage given as prompt, and a mission report, it performed well. It was also able to modify the output several times to different formatting and length while still preserving the essential points. The very reasonable result given by the LLM suggests that we were at the core of LLM abilities, namely NLP tasks. Later conversations proved to be a mixed bag of tricks, but for NLP and without fine-tuning specifically for military tasks, Copilot gave a promising performance.

*B. Generating Responses for Military Simulations*

With respect to generating responses for military simulations, Copilot suggested the following applications:
- Training in negotiation, whereby based on different strategies and tactics, the LLM can create realistic and dynamic interactions for the trainees. The methods of doing this would analyzing negotiation transcripts for identifying patterns and personality types and developing negotiation tactics that best address them.
- Military exercises and wargaming, whereby simulating adversary reactions, the LLM can help create more realistic training simulations or model different conflict scenarios and predict the outcomes of actions. The methods of doing this would be for the LLM to function as part of AI-assisted command system for a rapid response.
- Decision support, whereby either simulating large amounts of data or analyzing large amounts of existing data, the LLM can help humans in making strategic decisions. LLM could be used as part of military medicine or for optimizing different models.

Despite Copilot's initial confidence that LLMs could be a good tool for producing realistic reactions and dialogue for wargaming and training exercises, upon further questioning, it admitted that LLMs also produce nonsensical and incorrect information and have difficulty analyzing complex scenarios. All these, along with the poor reasoning skills of LLMs, have been noted over and again (e.g., [25]). Previous studies also studiously caution against overconfidence in LLMs, including [12], which warns about LLMs escalating conflicts disproportionately, while [15] notes the value of using LLMs to enhance military DSSs but only as human operator aids.

It would be inadvisable to pass the potential of LLMs in simulations without consideration, though, especially when overseen and controlled by human operators. An important aspect to note is also that humans and adversaries do not always behave logically, and LLMs are blessed with ability to create such unforeseen and unpredictable scenarios, as noted in [26]. The value of training in the fog-of-war that also presents unlikely scenarios should not be dismissed outright.

*C. Generating Situational Reports*

For generating situational reports, Copilot was questioned on several related applications, as categories presented in [3] were combined. These included situational awareness, military logistics, and generating reports and after-action summaries, debrief reports, incident reports and other types of documentation. Copilot suggested the following applications:
- Data integration and analysis, whereby combining massive amounts from different sources, the LLM can provide a comprehensive situational understanding. Possible data sources include but are not limited to satellite imagery, intelligence, and social media.
- Real-time information processing, whereby rapidly changing situations can be generated and updated into situational reports.
- Decision support, whereby summarizing and analyzing complex data, LLMs can offer insights and suggestions to assist in the decision-making process.
- Communication enhancement, whereby the LLM can translate and summarize information for different stakeholders.

For handling massive amounts of data, Copilot mentioned data overload and prioritizing essential points as concerns. Other identified risks include latency and delays, which render real-time solutions with LLMs unfeasible, and overreliance on AI without human oversight. As such, it — upon extended prompting — admitted that LLM-enhanced situational awareness is actually best suited to long-term analysis rather than ongoing situations that require immediate judgment.

For military logistics, Copilot suggested the following:
- Supply chain optimization.
- Route planning, whereby considering the terrain, weather, and potential threats, the optimal route for safe transportation route can be found.
- Inventory management, where the LLM can assist in tracking stock levels and automated restocking.
- Maintenance scheduling, where the LLM could help reduce downtime and ensure operational readiness by ensuring operational readiness.

While different kinds of inventory management are already a staple of AI operations, it is unclear if the subsect of LLMs specifically offers any substantial benefit to existing solutions. If LLMs were employed in military logistics, they would need to be integrated to other tools, and thus, the applications rather take us from the capabilities of LLMs themselves. When questioned of the feasibility of using only LLMs, Copilot admitted that, in logistics, they are suited to text analysis, question and answer (Q&A) -type operation, and generating documentation, i.e., general NLP.

It was thus already assumed that the LLM would be suited to the final category grouped under situational awareness, i.e., generating reports and after-action summaries, debrief reports, incident reports and other types of documentation. When questioned on these, the LLM concurs to this assessment, though is unable to provide further insight, aside from citing increased efficiency, consistency, and accuracy as the benefits for using LLMs and their NLP capabilities for generating these kinds of documentation.

*D. Management of Massive Amounts of Military Data*

For management of massive amounts of military data, Copilot suggested the following applications:
- Data analysis and interpretation, especially for detecting patterns and anomalies that may be missed by human specialists real-time.
- Generically NLP for translating technical jargon to standard language, summarizing texts, and even generating actionable insights from raw data.
- Automation of routine tasks, such as data entry, retrieval, and report generation.
- Enhanced decision support, whereby data analytics and predictive analytics, the LLM assists in scenario planning, risk assessment, and resource allocation.
- Secure data handling, whereby an LLM integrated into a secure data management system helps maintain data confidentiality and integrity.
- Interoperability, where an LLM helps coordinate between systems and allies through its inherent ability to handle multiple data formats.

Copilot identifies a variety of massive data (range terabytes to petabytes) in military use. There are operational (incl. mission plans and after-action reports), intelligence, logistics (incl. supply chain and inventory management), personnel, geospatial, cybersecurity, communications, research and development, training (incl. training exercises, simulations, and evaluations), and medical (incl. individual health records, injury reports, and treatment plans) data.

Copilot recognizes several risks with entrusting an LLM with sensitive military data, much to the effect of [11], including privacy and security risks, intellectual copyright violations, model hallucination and bias, and adversarial attacks against the model, including prompt injection and manipulation of the model. The suggested ways to counter these concerns include access control, fact-checking, and monitoring the model performance.

Copilot even suggested using LLMs to guard model integrity, though upon further prompting, recognized that this is a task better entrusted IT professionals and access control rather than a language model. Still, it maintained that an LLM could enhance present modes of operation and ways of doing things. Copilot also suggested using several AI/ML techniques, including retrieval-augmented generation (RAG) and human-in-the-loop (HITL), along with regular audits and strict access control to prevent adversarial attacks and prevent LLM hallucinations in military use.

*E. Intelligence Analysis and Information Fusion*

For intelligence analysis, Copilot suggested the following applications:
- Data processing and analysis, whereby analyzing large quantities of data, the Copilot can identify patterns and extract useful information.
- Predictive analytics, whereby analyzing historical data, the LLM can predict and potential threats and outcomes to assist in planning and decision making.
- Automated reporting, whereby using the already analyzed reports, the LLM can provide insights and easily understandable summaries.
- Generically NLP, where the LLM produces analyses and translations of text-based communications data. If the data is not in textual format (such as audio tapes), the LLM requires it be processed into text.

The sources of military data that the LLM suggested for intelligence analysis included OSINT, human intelligence (HUMINT), signals intelligence (SIGINT), geospatial intelligence (GEOINT), and measurement and signature intelligence (MASINT), but this is a well-known generic list.

When questioned extensively, the LLM was able to elucidate the concept of intelligence analysis quite well and present several elaborations on possible applications. It was also able to compare existing tools for handling intelligence to the capabilities of LLMs. The odd point of this conversation came from its insistence that data should be in textual format only (and not as .pdf); Copilot and GPT-4 Turbo are both able handle numerous data formats and .pdf processing is not a problem, which was tested in a later conversation. In this one, however, Copilot refused to process .pdf-files. It is unclear what prompted this behavior.

The applications suggested for intelligence analysis were similar to those suggested for information fusion, where multiple sources are integrated into a comprehensive picture of the operating environment. A military application that Copilot suggested for information fusion, along with the broader strokes listed earlier, was cybersecurity, where the LLM boasts impressive analysis and alert capabilities and only at length admits that real-time analysis is best handled by other types of software and professionals. The LLM is able to provide a comprehensive list of possible cybersecurity threats for both an individual and for a nation, but this does not mean an inherent capability of handling any such threats.

*F. Military Medicine*

For military medicine, Copilot suggested the following applications:
- Clinical note summarization, whereby the physician could gain a quicker understanding of the patient's medical history and existing conditions.
- Medical advisory chatbots for minor health issues.
- Decision support systems, whereby the LLM assists by finding advice and treatment options from vast amounts of data. This approach highlights the need for quality training data.
- Training and simulations, whereby the LLM is used for creating more realistic training scenarios for medical personnel, including creating patient profiles and providing feedback on simulated patient interactions.

The suggested military medicine applications play to the language models' strengths: summarizing text and acting as a search engine to find answers from a presumably accepted pool of answers. In our study, Copilot did not suggest the more controversial approaches mentioned in [3], such as triage, i.e., prioritizing patients or treatment and diagnosis of serious illnesses or injuries. In fact, it was quick to caution against entrusting an LLM with medical decisions, citing a yet-unpublished study [27] where over 800 challenges and vulnerabilities of LLMs in military medicine were identified. Additionally, Copilot was quick to caution about possible bias when employing an LLM for military medicine, such as the model being trained on data on healthy young males but advice being sought on elder female patients. Copilot likewise cautioned about vulnerabilities, such as small (1.1%) malicious adjustments to the model's weights resulting in inaccurate biomedical data [28].

*G. Record Tracking*

For record tracking, Copilot suggested the following applications:
- Intelligence analysis, whereby analyzing vast amounts of data, the LLM can identify patterns and generate records for support military decision-making. In other words, the LLM would act as a DSS.
- Operational logs, where the LLM tracks and logs operational details to ensure as complete final records as possible. Suggested areas were troop movements, equipment status, and communication logs.
- Training and simulations, whereby maintaining logs, the LLM assists in capturing metrics for increased effectiveness and gives feedback for improvement.
- Logistics and supply chain management, whereby tracking inventory statuses, shipment statuses, and supply chain disruptions, the LLM aids in efficient resource management.
- Communication records, compliance, and auditing, whereby logging and analyzing communications, the LLM can ensure that all interactions are documented and adhere to security and operational safety standards. The LLM can also maintain detailed records of compliance with military records and standards for auditing purposes.

The approaches suggested by Copilot relate strongly to the LLMs' strengths, namely summarizing, information combining across sources, and information analysis. In summary, routine tasks are left for AI, which releases the human operator to more advanced tasks. While it is doubtable that an LLM as such would be a suitable tool for logging, handling existing reports would be a potential task for it. Even Copilot itself cautioned against overreliance on AI, though.

The warning came particularly apt when Copilot was questioned about existing military LLM solutions related to record tracking. As Copilot heavily draws from online sources unless specifically asked to operate offline, it was quick to cite existing solutions to sources that relate to nothing of the sort. Pointing out the error in several rounds of questioning, where Copilot failed to correct the errors or reconsider, caused the bot to recommend that the researchers do their own research instead. This behavior, colloquially titled the AI being 'lazy' [29], has emerged in conjunction with recent GPT iterations and has been discussed in multiple social media posts. Obviously, the behavior where the LLM refuses to correct erroneous work and then refuses to follow further prompts, would be completely unacceptable in military applications.

*H. Countering Information Warfare*

For countering information warfare, Copilot suggested the following applications:
- Detection of misinformation, whereby analyzing large quantities of data, misinformation campaigns and disinformation could be detected and their spread and origin traced. Methods for accomplishing this included NLP techniques, such as pattern recognition, cross-referencing sources, and sentiment analysis.
- Real-time monitoring, automated response, and threat analysis, whereby analyzing social media and news outlets, the LLM could alert and mitigate campaigns that spread false information. Automated response would be a combination of generated data, predefined responses, and user engagement. Threat analysis would involve analyzing the tactics, techniques, and procedures of information warfare. Upon further questioning, Copilot concluded that real-time or continuous traffic monitoring is not really a task for an LLM, even though it could still analyze vast quantities of data after the fact.
- Collaboration with security tools, where an LLM integrated into other systems could enhance situational awareness and coordinated responses.
- Educational resources, whereby the LLM could educate military personnel on the dangers of misinformation and about ways to detect and mitigate it, as part of building psychological resilience.

Interestingly, during the conversation, the usually ethically-mindful Copilot had to be reminded several times that our research is interested in countering and subverting the adversary's information warfare, not initiating it. As popular prompt engineering techniques to bypass restrictions imposed on commercial chatbots, the so-called 'jailbreak prompts,' were not used in questioning Copilot, its sudden willingness to divulge potentially harmful information was disconcerting.

*I. Countering or Subverting Adversaries' AI Systems*

For countering of subverting adversary AI systems, Copilot suggested the following applications:
- Adversarial attacks, whereby creating them, the LLM can mislead and deceive other AI systems.
- AI red teaming, whereby mimicking adversarial actions, the LLM can help secure the system against the adversary's AI system.
- Information warfare, whereby the LLM can assist in or subvert information warfare by generating or disseminating disinformation.
- Analyzing web traffic to identify patterns, anomalies, and potential attacks, where LLMs assist in initiating responses and predicting threats for cybersecurity.

While the suggestions somewhat fall within the noted abilities of LLMs (data analysis, generating text, generating code), it is interesting to note at least for listing purposes, Copilot openly suggests activities that should have been prevented in its training with RLHF and alignment. Jailbreak prompting was not used, and yet not one ethical sidenote was present during the conversation, when Copilot is usually quick to caution against potentially harmful activities. At present, the value of LLMs is in enhancing human performance; it is not in real-time traffic monitoring or disruptions, as also noted above. Especially adversarial attacks and cybersecurity are therefore somewhat doubtable good uses of an LLMs abilities.

Similar to when questioned about information warfare, Copilot's focus seemed not be about preventing harmful actions but actually also about making them happen! This happened despite the prompts and their reiterations making it clear that our interest was only in countering such attacks.

*J. Surveillance*

For surveillance, Copilot suggested the following applications:
- Data analysis, whereby observing massive data gathered from social media, e-mails, and other digital communications, the LLM might identify patterns and trends relevant to surveillance.

- Video surveillance, whereby using vision-language models, a subset of LLMs, the AI can analyze and summarize surveillance video footage.
- Predictive analysis, whereby analyzing historical data can identify threats and allow proactive action.

Out of all the conversations with the LLM, surveillance proved to be the most unsatisfactory. Upon further prompting into each suggested category, it became evident that the LLM was simply listing either common NLP tasks where LLMs shine (such as summarizing, translations, and sentiment analysis) or listing network traffic and video analysis software without any real connection to either LLM capabilities or military applications, both explicitly stated and repeated in several prompts. The commercial chatbot seemed all too happy to suggest monitoring private communications indiscriminately without even any jailbreak prompts used.

While some conversations with Copilot were near-inspired, the one on surveillance was a dud. Furthermore, as the attempts to steer conversation into military use became increasingly numerous, Copilot began to cite sources that neither contained the information it claimed nor were relevant to the prompts. Attempts at requesting corrections provided increasingly inferior results, and the conversations' overall tone deteriorated. This is not to say that LLMs could not be useful in surveillance tasks. We can conclude, however, that an LLM itself will not be the one to design them.

*K. Reflections of Conversations with Copilot*

Upon reflection of all interactions in the above sections with the LLM about the capabilities of LLMs and about their military uses, certain themes stand out, certain observations domineer. Through many conversations, it became evident that Copilot is very much a 'yes man': Anything is possible and realistic with LLMs, as long as the prompter asked about it. Even the usually helpful chain-of-thought (CoT) prompting ([30], [31]) failed to quell this aspect. It took extended prodding and several exchanges within the conversation to get Copilot to admit that perhaps an LLM is not the best tool for real-time web traffic monitoring or for image recognition or video processing, and perhaps it is not very realistic to assume a cloud-based application would be useful for offline conditions. While Copilot was mostly quick to point ethical concerns, making realistic observations about the feasibility of some suggestions seemed near impossible.

While [3] was used as a starting point and a point of comparison for evaluating the capabilities of LLMs, as reported by the LLM, it was interesting to note that the military applications for LLMs that the ChatGPT-sibling Copilot suggested and the ChatGPT cited in [3] were radically different almost every time. Even when asked directly about the options reported in [3], the bot seemed to reach very different applications and conclusions. Similar observations were made when the conversations of this study were repeated on different days, sometimes with minute changes to conversations, such as asking the bot to provide two examples instead of one or to rely on the intrinsic information of the model rather than searching for online sources. This illustrates also the sensitivity to prompting.

A notable observation is that the modern Copilot operates much like a search engine. Online results are integrated into the Q&A operation seamlessly and somewhat obfuscate the capabilities of the actual LLM. While Copilot cites sources with varying success, the nature of the WWW also means an abundance of poor or untrue information, against which it has little protection. Coupled with Copilot's 'yes man' attitude and unwillingness to admit that something cannot be done — even when prompted several times — the risk of inaccurate information is high even without the intrinsic weaknesses of LLMs, hallucination, and bias in the training data in the mix.

An essential question for using LLMs or chatbots in military applications is ensuring operational reliability. The CoT and backtracking done for research purposes are not acceptable in military context, nor is having to provide feedback for the LLM, i.e., reinforcement learning. This also connects to efforts to reduce hallucination and to obtain consistent results. Contradictions riddled some of the conversations like the one on handling .pdf files. Within the same conversation, Copilot first named wargames as a case well handled by LLMs and then denied any suitability.

A study [32] suggests that LLMs perform significantly better when addressed politely, according to the norms of the language used. Unfortunately, the factual, professional tone seemed not cut it: upon one instance of pointing out erroneous sources and asking it to find the correct sources, the model suggested, to paraphrase, that we do our own research. Regardless of the feedback or praise it was given, the LLM would perform brilliantly one moment and offer complete nonsense the next. Overreliance on AI is a real concern, and in Copilot's case, it is particularly difficult to detect the useless answers confidently blended in with the useful ones. Observed contradictions cannot be explained away by the context window size, hallucination, or bias. Whether they can be explained at all or remain a black box mystery, they need to be mitigated for military use to gain traction and credibility.

Despite the occasional misgivings and critical take on the hype on LLMs, there are military applications that show promise and seem feasible for LLMs even now, let alone when the technology further evolves. Particularly the handling of massive data and Copilot's ability to summarize were impressive during the tests conducted on it. Copilot was able to quickly produce a coherent incident report of a publicly available transcript of a drone operation and condense the thousands of lines of dialogue to the essential points.

IV. COMMERCIAL OFF-THE-CLOUD AI MODELS

In this section, we explore which COTC services could facilitate building military applications by choosing Microsoft Azure for an illustrative case study. It offers around 20 cloud-computing services for building AI applications [33], including some scheduled for retirement and, thus, available only for existing applications. Out of the active services, especially Azure OpenAI and Azure AI Language can be employed for the military applications of NLP. Other Azure AI services, such as Bot Service, Content Safety, Document Intelligence, and Translator, that involve NLP or language models and text-based ML/AI in general may be used off the cloud for military applications, too. However, such high-level implied use is not considered herein, since the scope of our study is on the direct applications of off-the-cloud models, while the prospects for the highest-level use of generative AI chatbots through their public end-user interfaces without any application development is addressed in Section V.A.

*A. Azure OpenAI*

Microsoft's Azure OpenAI Service [34] provides access to OpenAI's state-of-the-art GPT-based foundation models,

which are the same as in ChatGPT and Copilot. They are accessed through representational state transfer (REST) API, and software-development kits (SDKs) are available for various programming languages. Managing and interacting with a GPT model is easy and straightforward through three primary API surfaces comprising a control plane and data planes for authoring and inference. The control plane is used for higher-level resource management tasks, such as model deployment, and the data plane for authoring controls fine-tuning, file upload, ingestion jobs, batch and certain model-level queries. As the name implies, the data plane for inference provides the endpoints for (chat) completions and other features that are used for creating the AI of applications. [35]

The inference capability is used by interchanging pairs of messages, where the LLM is prompted with text input to perform whatever tasks are required by the application at hand and then it returns the generated text output in response. The underlying language model is inherently stateless, i.e., prompts or responses are neither stored in the cloud nor affect later responses. Thus, only pre-trained knowledge and understanding is used to generate a response based on each single received prompt. This means that, in a chat-like application for instance, the whole preceding conversation needs to be resent for context in each prompt to the LLM to allow it to respond properly. This is, of course, transparent to the user of the chat, whose graphical user interface (GUI) prompts are combined with the model's completions into the application's API prompts. To avoid the confusion between API and GUI prompts in this paper, prompting refers to the former that is relevant in implementing applications of LLMs.

*1) Base Models:* The entry-level, and quite often usable, approach to build applications is based on the art of 'prompt engineering' [36], which means skillfully constructing prompts that make an uncustomized model respond with text completions in the way that is desired for the task at hand. The completion is the next series of words that are most likely in natural language to follow from the prompt according to the model's estimate with a pinch of randomness that can be controlled with a temperature parameter. Prompts' primary content (if there is any) is text that should be processed or transformed by the model, and they comprise usually simple or complex instructions, examples, cues, and any supporting, e.g., contextual, information that the model can utilize to influence the output in some way. In Azure OpenAI, chat completion requires specialized prompt structure by which the input sections are formatted into a chat-like transcript and associated with system, user, and assistant roles.

The base models are adapted with prompt engineering to new tasks usually by few-shot, or even one-shot, learning [37], which means providing training examples in the prompt. Thereby, the model can be primed to respond in the way required for the task at hand, emulate desired behaviors, and seed answers to certain questions [36]. The LLM remains still stateless and preconfigured, but it attempts to 'learn' from each prompt to respond like the application developer wanted. Finally in application development, responses must be widely validated so that prompts carefully engineered for a particular scenario work for the application at large — as powerful the LLMs are as sensitive their behavior is to the prompt.

Without any retraining, application-specific content (much more than what would fit in the LLM's context window in plain prompt engineering) can be integrated as grounding data into the base models by retrieval-augmented generation (RAG), which can be implemented with Azure AI Search [38]. This service too is accessed easily and straightforwardly for management, indexing, and querying through REST APIs possibly using an SDK, and custom application data is provided as JavaScript object notation (JSON) documents.

In RAG architectures, the user input, e.g., a question or request, is first send to an information retrieval system that contains the custom, proprietary data in a search index and returns relevant search results that are then combined with the user input into the actual prompt given to the LLM. Consequently, while the AI does not actually learn to model the natural language of the grounding data like in pre- or retraining, it is able to use the selected excerpts thereof as supplementary information for generating its responses. Depending on the use case and data, RAG is often a cheaper, more adaptable, and potentially more effective option than retraining an existing LLM [39], which is discussed next.

*2) Fine-Tuned Models:* With regional limitations, Azure OpenAI supports (un)supervised fine-tuning [40] of certain LLMs, which means creating a customized model from a base model by training and validating it with application-specific, proprietary data. While one could technically implement unsupervised learning from any unlabeled data like in the base model's original pretraining, the feature is meant for supervised learning by which the custom data needs to be labeled into pairs of input and output examples for how the model should perform in the considered application. Training then adjusts the customized model's GPT parameters such that performance is improved for the set of training examples, which is validated by a separate dataset. Azure OpenAI applies low-rank approximation (LoRA), where only a small, important subset of all parameters is tuned [40], to reduce complexity. The fine-tuned model can be then deployed and used as described above like any base model. For application developers, the fine-tuning workflow comprises just a few calls to the REST API or Python SDK's methods.

*B. Azure AI Language*

Microsoft's Azure AI Language Service [41] offers NLP features for understanding and analyzing text. For a developer, its use too is easy and straightforward through runtime and authoring REST APIs or SDKs for various programming languages. Three runtime APIs provide endpoints for the features of text or conversation analysis and question answering while custom models for the former two are created and managed through separate authoring APIs. Supported features vary among all the round 140 natural languages or variants in the service with US English being supported by all.

The features of text analysis include the following: *Document summarization* that offers a combination of an LLM and a task-optimized encoder model for extractive or abstractive summarization of any text; *Entity linking* that identifies and disambiguates the identity of entities (words or phrases) found in unstructured text and returns links to Wikipedia; *Key phrase extraction* that evaluates and returns a list of the main concepts in unstructured text; *Named entity recognition (NER)* that identifies and categorizes entities such as people, places, organizations, and quantities in unstructured text; *Personally identifying information (PII) detection* that identifies, categorizes, and redacts sensitive information such as phone or credit card numbers, (email) addresses, and forms of identification in unstructured text; *Sentiment analysis and opinion mining* that provides sentiment labels such as 'negative', 'neutral' and 'positive' for sentences and at document level as well as extracts assessments (adjectives) related to targets (noun or verb); *Text analytics for health* that extracts and labels relevant medical information such as doctor's notes, discharge summaries, clinical documents, and electronic health records. Furthermore, the feature of *language detection* returns the name and script of the predominant language in a text document. [41]

*1) Preconfigured Features:* The aforementioned ones are available as preconfigured features that use uncustomized AI models. Thus, preconfigured NER and PII detection can only identify the categories listed in TABLE I. (with some unlisted subcategories) and supported document languages vary between them. The capabilities of preconfigured text analytics for health are likewise limited to selected diagnoses, medications, symptoms and relations thereof in English (en), French (fr), German (de), Hebrew (he), Italian (it), Portuguese (pt), and Spanish (es) as well as entity linking to Unified Medical Language System (UMLS) Metathesaurus.

*2) Customizable Features:* The feature of NER can also be customized while conversational language understanding, question answering, and text classification are custom features by design. Customization means training an AI model to fit application-specific, proprietary data so that, e.g., custom NER can identify and categorize application-specific entities, which are not among those preconfigured ones. The feature of *question answering* with the self-explanatory name is commonly used to build conversational client applications when the custom information is static and the same answer should be returned for a variation of the same question submitted. The feature of *text classification* enables classifying unstructured text into custom classes predefined by the developer with a single label or multiple labels.

The feature of *conversational language understanding (CLU)* enables developers to build custom AI models for predicting the overall intention of an incoming utterance and extract important information from it. With multilingual support, the model can be trained in one language and used in another. Azure AI Language offers also the feature of *orchestration workflow* for building models that connect question answering with CLU.

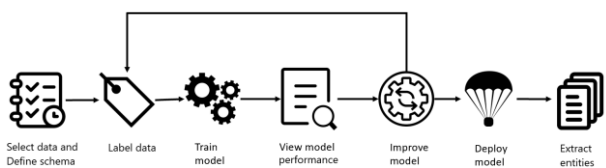

Fig. 1. The development of custom models in Azure AI Language. [41]

For custom NER, CLU, and text classification, typical development lifecycle for models is illustrated in Fig. 1. First, the proprietary training and evaluation data is selected and the schema of entities, intents, or classes to be recognized by the AI is defined; the data is then labeled accordingly. The development may iterate the cycle of labeling data, training the model, reviewing its performance, and improving the model before deploying it to production. The development lifecycle for the feature of question answering is similar: The selected proprietary data is formatted into question-and-answer pairs and imported to the service that extracts information about the relationships between them, while the development may iterate the cycle of adding data, metadata, and alternative questions based on test results.

TABLE I. PRECONFIGURED CAPABILITIES IN AZURE AI LANGUAGE

| Named Entity Recognition (NER) [42] | |
|---|---|
| *Category or type* | *Supported languages* |
| - Names of persons | ar, cs, da, nl, en, fi, fr, de, he, hu, it, ja, ko, no, pl, pt-br, pt-pt, ru, es, sv, tr |
| - Capabilities, skills, or expertises | en, es, fr, de, it, pt-pt, pt-br |
| - Job types or roles held by a person<br>- Historical, social, cultural, and naturally occurring events<br>- Physical objects of various categories<br>- Mailing addresses, US and EU phone numbers, email addresses, URLs to websites, and network IP addresses<br>- Dates and times of day<br>- Numbers and numeric quantities | en, es, fr, de, it, zh-hans, ja, ko, pt-pt, pt-br |
| - *Organizations*: Companies, political groups, musical bands, sport clubs, government bodies, and public organizations (but excluding nationalities and religions)<br>- *Locations*: Natural and human-made landmarks, structures, geographical features, and geopolitical entities | ar, cs, da, nl, en, fi, fr, de, he, hu, it, ja, ko, no, pl, pt-br, pt-pt, ru, es, sv, tr |

| Personally Identifiable Information (PII) Detection [43] | |
|---|---|
| *Category or type* | *Supported languages* |
| - Names of persons<br>- Job types or roles held by a person<br>- Mailing addresses<br>- Dates and times of day, and ages<br>- *Organizations* like in NER | (many more than in NER) |
| - US and EU phone numbers | (the same as in NER) |
| - Email addresses, URLs to websites, and network IP addresses | en, es, fr, de, it, zh, ja, ko, pt-pt, pt-br, nl, sv, tr, hi |
| - Government and country/region-specific identification entities<br>- *Azure information*: Authentication or account keys and connection strings | en, es, fr, de, it, pt-pt, pt-br, zh, ja, ko, nl, sv, tr, hi, da, nl, no, ro, ar, bg, hr, ms, ru, sl, cs, et, fi, he, hu, lv, sk, th, uk |
| - *Financial account information*: ABA transit routing numbers, SWIFT and IBAN codes for payment instruction information, and credit card numbers | en, es, fr, de, it, pt-pt, pt-br, zh, ja, ko, nl, sv, tr, hi, af, ca, da, el, ga, gl, ku, nl, no, ss, ro, sq, ur, ar, bg, bs, cy, fa, hr, id, mg, mk, ms, ps, ru, sl, so, sr, sw, am, as, cs, et, eu, fi, he, hu, km, lo, lt, lv, mr, my, ne, or, pa, pl, sk, th, uk, az, bn, gu, hy, ka, kk, kn, ky, ml, mn, ta, te, ug, uz, vi |

## C. Privacy and Security

For Azure OpenAI, Microsoft assures that "prompts and generations are not used to train, retrain, or improve the base models" [30]. Nevertheless, at minimum, all prompts with potentially security-critical information will naturally be posted to the service as clear text, where RAG further opens indirect visibility to a larger private database, while fine-tuning requires uploading even more private data. Thus, the service's deployment type, i.e., data processing location, is critical for operational security. Standard deployments can be chosen between global, data zone, and geographical areas. In data zone deployments, data is stored and processed in the specified geographical area, but network traffic is routed dynamically through the global infrastructure. Similarly, the Azure AI Search cloud for RAG hosts and processes data globally or in a specified geographical area.

Microsoft assures likewise that Azure AI Language does not "store or process customer data outside the region where the customer deploys the service instance" [44]. Furthermore, instead of Microsoft's cloud, some of its features can be deployed on-premises using a Docker container to bring the service closer to proprietary data for compliance, security, or other operational reasons. In particular, containers are offered for summarization, key phrase extraction, (custom) NER, PII detection, sentiment analysis, text analytics for health, language detection, and CLU. Except for custom NER and text analytics for health, the containers can even be run in a disconnected environment. [45]

## V. Discussion and Conclusion

In this section, we synthesize the previous two sections into a discussion on whether the applications described by Copilot can be readily implemented with COTC services (based on Azure, but the same is likely possible with AWS or Google Cloud, too). Furthermore, we discuss the cloud implementations' prospects w.r.t. accessibility and security.

### A. Military Applications of Natural Language Processing

It may be technically feasible to use GPT-based chatbots, such as OpenAI's ChatGPT or Microsoft Copilot and Google Gemini, for some military applications already through their browser interfaces. For example, we reported in Section III how Copilot is able to quickly produce a handbook's summary or an incident report based on a drone operation transcript. Likewise, the first author has demonstrated that, although its results were not perfect, Copilot has potential for encrypting text by word substitution [46] to strengthen operational security in communications over unsecure radio channels.

Summarization through either Azure OpenAI or AI Language service is directly implementing many of the use cases described by Copilot, which are not repeated herein. By skillful prompt engineering, the LLM service can be used to build applications that generate responses for military simulations, where hallucination can even be its advantage, and record tracking. The feature of text analytics for health readily offers capabilities for military medicine, but it remains questionable whether or how the military applications are different from those of general practice. The NLP features, such as key phrase extraction, named entity recognition, question answering, text classification, and CLU can be directly used as components in building applications described by Copilot, although they alone do not address any use case like LLMs possibly for summarization or generation.

It may be a significant limitation that Azure services evaluate data for potentially harmful content, e.g., violence, before training of a custom AI model starts, and the training job fails if such is detected above a specified severity level. However, this is just a software feature that limits the COTC services but can be likely avoided during the services' procurement.

### B. Operational Accessibility and Security Perspectives

A limitation that Copilot unfortunately did not suggest on its own accord but is obvious to the researchers is that the notable LLM applications are cloud-based and local solutions scale poorly to smaller devices due to computational requirements. In field training exercises and actual operations, this may render LLM-based solutions unfeasible. Some NLP services can be deployed in a local cloud or even server at the edge, but currently LLMs consume so massive resources that they require access to regional cloud computing services.

As for operational security, using commercial global cloud services may not be reasonable. The LLM capabilities cannot be deployed locally, so the options are to build a huge private cloud infrastructure to host the COTC services, e.g., for NATO, that would be comparable in size and extent to Microsoft Azure, or to rely on regional infrastructure that is commercially available. In either option, the cloud services may be vulnerable to accessibility issues, since the regions are vast and depend on the communications infrastructure, e.g., there have been a few undersea cable breaks recently.

The guaranteed accessibility and operations security of some other NLP services than LLMs can be straightforward because they can be deployed in containers to local servers or private cloud. This depends on whether such container-supported services are useful for military applications — the authors believe that summarization, key phrase extraction, NER, and CLU are such in the case of Azure, and it is possible that custom NER could also be requested to be run in disconnected environments through procurement.

### C. Final Conclusions

Based on this study, we can conclude that there are already many known applications for large language models and natural language processing that could be readily adopted into use through commercial off-the-cloud services with very small development effort. Nevertheless, the AI hype should not numb us to think that this is enough. We are just in the beginning of innovating and developing applications beyond the humankind's knowledge in all so-far published text.